\documentclass{article}
\usepackage{spconf,amsmath,graphicx}

\usepackage[T1]{fontenc}
\usepackage{caption}

\usepackage{enumitem}
\setlist{nosep, leftmargin=14pt}

\usepackage{graphicx}
\usepackage{amsmath}
\usepackage{amssymb}
\usepackage{booktabs}
\usepackage{pifont}
\usepackage{multirow}
\usepackage{soul}
\usepackage[capitalize]{cleveref}

\usepackage[table,dvipsnames]{xcolor}
\newcommand{\slip}{\mathrm{SLIP}}
\DeclareMathOperator*{\argmax}{arg\,max}

\newcommand{\std}[1]{\textcolor{gray}{\scriptsize\,±#1}}

\title{Slide-Level Prompt Learning with Vision Language Models for Few-Shot Multiple Instance Learning in Histopathology}

\name{
\begin{tabular}{c}
Devavrat Tomar$^{\star}$ \qquad Guillaume Vray$^{\star}$ \qquad Dwarikanath Mahapatra$^{\dagger}$ \qquad Sudipta Roy$^{\ddagger}$ \\
Jean-Philippe Thiran$^{\star}$ \qquad Behzad Bozorgtabar$^{\star}$
\end{tabular}
}

\address{$^{\star}$ LTS5, EPFL, Switzerland \\
         $^{\dagger}$ Monash Medical AI Group, Monash University, Australia;
         $^{\ddagger}$ Jio Institute, India}

\begin{document}
\maketitle
\begin{abstract}
   In this paper, we address the challenge of few-shot classification in histopathology whole slide images (WSIs) by utilizing foundational vision-language models (VLMs) and slide-level prompt learning. Given the gigapixel scale of WSIs, conventional multiple instance learning (MIL) methods rely on aggregation functions to derive slide-level (bag-level) predictions from patch representations, which require extensive bag-level labels for training. In contrast, VLM-based approaches excel at aligning visual embeddings of patches with candidate class text prompts but lack essential pathological prior knowledge. Our method distinguishes itself by utilizing pathological prior knowledge from language models to identify crucial local tissue types (patches) for WSI classification, integrating this within a VLM-based MIL framework. Our approach effectively aligns patch images with tissue types, and we fine-tune our model via prompt learning using only a few labeled WSIs per category. Experimentation on real-world pathological WSI datasets and ablation studies highlight our method's superior performance over existing MIL- and VLM-based methods in few-shot WSI classification tasks. Our code is publicly available at \texttt{\color{magenta}{https://github.com/LTS5/SLIP}}.

\end{abstract}

\section{Introduction}
\label{sec:introduction_new}
Histopathological images are vital for cancer diagnosis and prognosis \cite{li2022comprehensive}, and their conversion into whole slide images (WSIs) allows for deep learning-based analysis. However, challenges arise due to WSIs' gigapixel resolution, requiring them to be divided into smaller patches for processing, and the limited availability of detailed annotations, which complicates supervised learning. As a result, WSI classification is typically approached as a multiple instance learning (MIL) task \cite{lu2021data} within a weakly supervised learning framework, yet these methods struggle with limited annotated data and under-representation of tumor types in public datasets.

Self-supervised learning (SSL) approaches \cite{stegmuller2024self,koohbanani2021self} help by learning general patch features, but developing a classifier still requires supervised training, which is challenging in few-shot settings. Vision-language models (VLMs) \cite{radford2021learning,zhang2023text,qu2024rise,stegmuller2025simple} offer an alternative, using natural language to guide zero-shot or few-shot settings by linking visual features with class-specific textual embeddings. These quantitative measures are then aggregated through a permutation-invariant operator or soft voting technique \cite{zhang2023text} to classify the WSI.

Our approach leverages large language models and a novel pooling strategy to align patch features with tissue-specific and WSI class embeddings in a MIL framework. Key contributions include:

\begin{itemize}
\item A novel MIL framework, built on VLMs,  utilizes the prior knowledge of large language model  to identify relevant tissue types for WSI classification;

\item  An efficient prompt-based learning strategy, $\slip$, ``\textbf{S}lide-\textbf{L}evel \textbf{I}ntegration for \textbf{P}rompt learning'', tailored for few-shot MIL WSI classification;

\item A new pooling strategy that aligns patch visual features with domain-specific tissue types as well as the alignment between tissue-specific embeddings and the whole-slide class descriptor, enhancing interpretability; 

\item Evaluation on two real-world hematoxylin and eosin (H\&E)-stained WSI datasets, demonstrating superior few-shot classification and generalizable representations over state-of-the-art MIL and VLM methods.
\end{itemize}

\section{Methods}
\label{sec:method}

\begin{figure*}[ht]
    \centering
    \includegraphics[width=\textwidth, height=0.48\textheight]{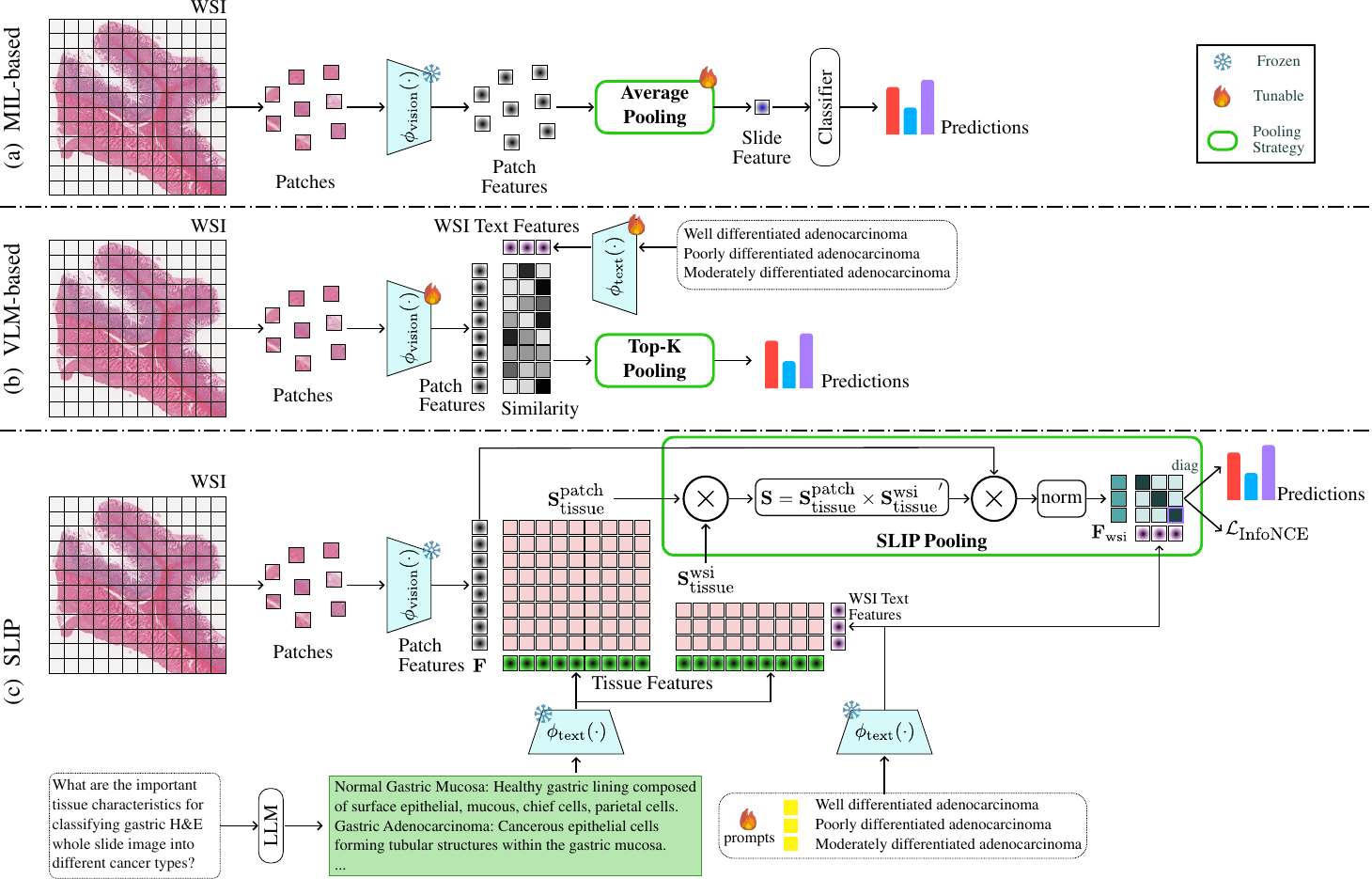}
    \caption{
    \textbf{Overview of the proposed method vs. existing MIL- and VLM-based approaches for few-shot WSI classification.} (a) Conventional MIL methods use pooling functions like Average Pooling for slide-level features; (b) VLM-based methods measure similarity between patches and WSI text prompts, using e.g., Top-K Pooling; (c) Our SLIP framework introduces SLIP pooling by computing similarity $\mathbf{S_\text{tissue}^\text{patch}}$ between patch features and tissue-specific text embeddings from ChatGPT, and $\mathbf{S_\text{tissue}^\text{wsi}}$ between whole-slide and tissue type names, aggregating class-specific features as $\mathbf{F}_\text{wsi}$.}
    \label{fig:enter-label}
\end{figure*}
 \noindent
We propose a framework that adapts VLMs for the few-shot classification of WSIs. Figure \ref{fig:enter-label} depicts the SLIP pipeline for slide-level prompt learning using a pre-trained VLM, distinguishing it from existing MIL and VLM-based methods.

\begin{table*}[t]
    \centering
    \caption{\textbf{Performance comparison on PatchGastric 3-category few-shot classification.} Best results per backbone are in \textbf{bold}, and second-best are \underline{underlined}. $^\ast$Results from \cite{zhang2023text} on the same split. Zero-shot follows general VLM settings, using WSI class names (excluding tissue types). Predictions are made per patch and averaged for WSI-level results.}
    \resizebox{\textwidth}{!}{
    \begin{tabular}{c|c|cccccc|cccccc}
    \toprule
       \multirow{2}{*}{Methodology} & \multirow{2}{*}{Baselines} & \multicolumn{6}{c|}{CLIP ViT-B/16} & \multicolumn{6}{c}{BiomedCLIP ViT-B/16} \\
       & & 1 & 2 & 4 & 8 & 16 & all & 1 & 2 & 4 & 8 & 16 & all \\
    \midrule
        \multirow{3}{*}{MIL-based} 
        & Linear-probe & 47.0\std{2.5} & 48.1\std{2.2} & 54.5\std{1.0} & \underline{64.1\std{0.7}} & \underline{66.2\std{1.1}} & 68.5\std{1.4} 
        & 47.9\std{2.5} & 55.1\std{1.9} & 59.4\std{1.1} & 66.3\std{0.7} & 62.1\std{4.0} & 67.7\std{1.5} \\
        & CLAM \cite{lu2021data} & 54.1\std{1.6} & 52.7\std{0.8} & 54.3\std{4.2} & 63.2\std{2.1} & 65.7\std{1.3} & 65.6\std{1.1} 
        & 49.6\std{3.7} & 55.2\std{2.4} & 59.1\std{2.4} & 66.8\std{2.1} & 62.2\std{2.4} & 68.3\std{1.3} \\
        & TransMIL \cite{shao2021transmil} & 43.5\std{7.7} & 42.7\std{3.0} & 54.4\std{3.2} & 61.6\std{1.2} & 63.7\std{1.3} & 66.5\std{1.3} 
        & 49.8\std{2.0} & 55.5\std{4.5} & 59.1\std{0.9} & \underline{66.8\std{1.4}} & \underline{66.1\std{1.1}} & 67.9\std{1.0} \\
        & PANTHER \cite{song2024morphological} & 45.0\std{1.7} & 43.3\std{6.4} & 49.3\std{1.7} & 60.1\std{1.0} & 63.0\std{1.7} & 67.2\std{1.8} 
        & 42.8\std{3.8} & 47.6\std{2.6} & 51.0\std{1.1} & 63.0\std{1.0} & 57.8\std{0.8} & 64.9\std{1.7} \\
    \cmidrule{1-14}
        \multirow{4}{*}{VLM-based} 
        & Zero-Shot & 52.3 & 52.3 & 52.3 & 52.3 & 52.3 & 52.3 
        & \underline{51.2} & 51.2 & 51.2 & 51.2 & 51.2 & 51.2 \\
        & CITE$^\ast$ \cite{zhang2023text} & \underline{60.1\std{0.9}} & \underline{59.0\std{0.1}} & \underline{60.9\std{0.9}} & 63.2\std{0.2} & 65.9\std{0.5} & \underline{68.7\std{0.6}} 
        & - & - & - & - & - & - \\
        & CoOp \cite{zhou2022learning} & 40.8\std{1.1} & 40.9\std{2.4} & 46.0\std{1.9} & 54.2\std{0.6} & 56.7\std{0.5} & 63.2\std{0.5} 
        & 40.3\std{0.5} & 40.8\std{0.7} & 48.6\std{0.5} & 57.5\std{0.6} & 61.0\std{0.8} & 65.1\std{0.4} \\
        & TOP \cite{qu2024rise} & 42.7\std{0.4} & 45.5\std{6.1} & 48.0\std{1.3} & 54.7\std{0.7} & 58.9\std{0.9} & 66.5\std{2.3} 
        & - & - & - & - & - & - \\
        \cmidrule{2-14}
        & SLIP-zero & 51.2 & 51.2 & 51.2 & 51.2 & 51.2 & 51.2 
        & 49.3 & 49.3 & 49.3 & 49.3 & 49.3 & 49.3 \\
        & SLIP-topk & 49.8\std{4.7} & 40.7\std{2.3} & 50.0\std{5.1} & 54.5\std{1.1} & 60.4\std{1.0} & 66.7\std{1.5} 
        & 51.1\std{1.3} & \underline{60.3\std{1.3}} & 59.1\std{0.7} & 61.0\std{2.3} & 59.1\std{1.0} & \textbf{70.9\std{0.5}} \\
        & SLIP-avg & 52.0\std{6.6} & 45.8\std{3.0} & 58.2\std{1.1} & 63.3\std{1.4} & 66.4\std{0.6} & 68.8\std{0.7} 
        & 48.7\std{1.2} & 59.0\std{1.7} & \underline{60.2\std{1.0}} & 66.7\std{0.4} & 59.1\std{0.7} & 68.0\std{1.0} \\
        \rowcolor{cyan!20}& \textbf{SLIP-our pooling} & \textbf{60.8\std{2.5}} & \textbf{61.7\std{3.9}} & \textbf{62.2\std{1.1}} & \textbf{65.0\std{0.8}} & \textbf{66.9\std{0.4}} & \textbf{70.0\std{0.6}} 
        & \textbf{51.4\std{2.2}} & \textbf{60.4\std{2.4}} & \textbf{61.7\std{1.2}} & \textbf{67.8\std{1.3}} & \textbf{68.3\std{1.5}} & \underline{70.6\std{0.6}} \\
    \bottomrule
    \end{tabular}}
    \label{tab:merged_gastric_results}
\end{table*}

\begin{figure*}[t]
    \centering
    \includegraphics[width=\linewidth, height=5cm]{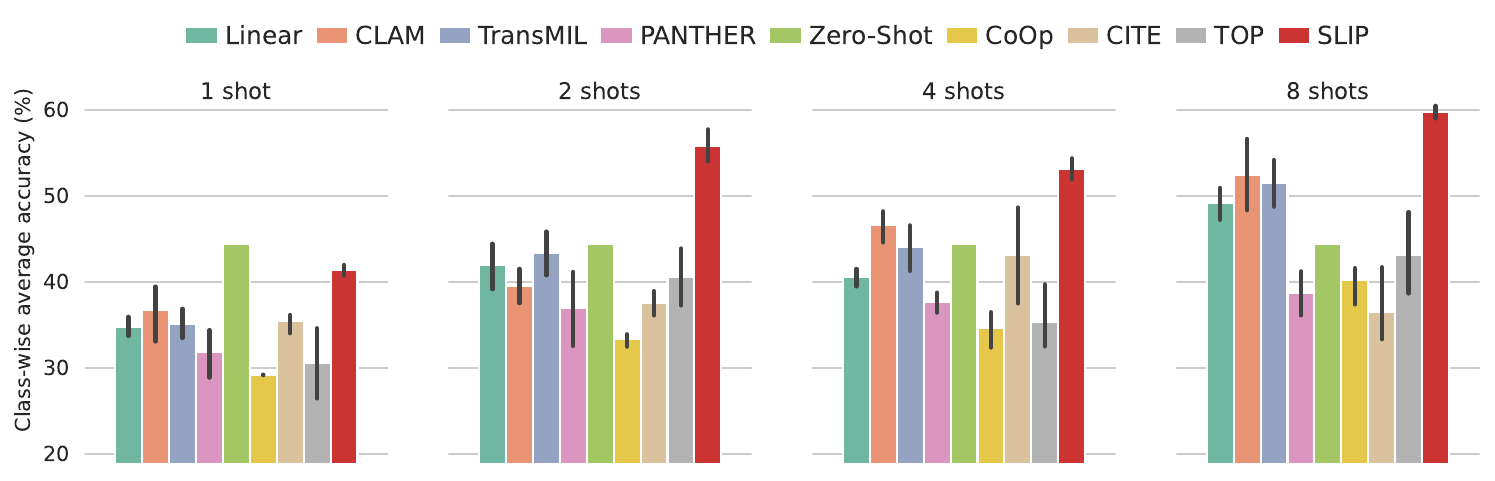}
    \caption{
    \textbf{Accuracy on the 3-category lung adenocarcinoma classification task (DHMC dataset)}. All baselines use the ViT-B/16 encoder from CLIP, with mean accuracy and standard deviation (error bars) shown across 10 runs.}
    \label{fig:dhmc}
\end{figure*}

\noindent
{\bf Notations.} Let $\phi_\text{vision}$ and $\phi_\text{text}$ denote the vision and text encoders of the pre-trained VLM, e.g., CLIP \cite{radford2021learning}. A whole slide image $X$ is represented as a collection (bag) of $N$ non-overlapping patch instances $X = \{x_1, x_2, ..., x_N\}$, and the textual class names relevant to the whole slide are denoted by $Y_\text{wsi}=\{y_\text{wsi}^1, ..., y_\text{wsi}^C \}$, where $C$ denotes number of whole-slide level categories. Additionally, textual descriptions of tissue types and characteristics specific to the domain of histopathology classification task, obtained by querying ChatGPT, are represented as $Y_\text{tissue}=\{y_\text{tissue}^1, ..., y_\text{tissue}^K\}$, where $K$  represents the number of tissue types sourced from ChatGPT.

\noindent
{\bf Quantifying tissue types similarities.}
Using ChatGPT \cite{openai2023gpt}, we generate descriptions of tissue phenotypes in WSIs as linguistic priors, adding domain-specific knowledge. For example, the prompt \textbf{\textit{“Provide a list of tissue types for identifying stages of adenocarcinoma in gastric histopathology whole slide images”}} yields detailed visual descriptors for subtyping gastric adenocarcinoma. At the outset of our approach, we compute the similarity between whole slide level class names, denoted as $Y_\text{wsi}$, and domain-specific tissue types $Y_\text{tissue}$, utilizing the text encoder $\phi_\text{text}$. This calculation is performed using softmax cosine similarity to produce $\mathbf{S_\text{tissue}^\text{wsi}}$, formulated as follows:
\begin{equation}
    \mathbf{S}_\text{tissue} ^ \text{wsi} (i, j) = \frac{\exp \big(\phi_\text{text}(y_\text{wsi} ^ i) \cdot \phi_\text{text}(y_\text{tissue}^j)/ \tau\big)}{\sum_j \exp\big(\phi_\text{text}(y_\text{wsi} ^ i) \cdot \phi_\text{text}(y_\text{tissue}^j) / \tau\big) }
    \label{Eq:1}
\end{equation}
where $\tau$ refers to the temperature parameter that regulates the sharpness of the similarity. The formula assesses the relevance of specific tissue types for classifying a whole slide image into class $y_\text{wsi}^i$, as indicated by $i^{th}$ row of  $\mathbf{S_\text{tissue}^\text{wsi}}$.
\par
Furthermore, we derive $\mathbf{S_\text{tissue}^\text{patch}}$ to quantify how well visual embeddings of patch instances align with textual embeddings of domain-specific tissue types, as specified by the following equation:
\begin{equation}
    \mathbf{S_\text{tissue}^\text{patch}}(n, j) = \frac{\exp \big(\phi_\text{vision}(x_n) \cdot \phi_\text{text}(y_\text{tissue}^j)/ \tau \big)}{\sum_j \exp\big(\phi_\text{vision}(x_n) \cdot \phi_\text{text}(y_\text{tissue}^j)/ \tau\big) }
\end{equation}
Leveraging both $\mathbf{S}_\text{tissue} ^ \text{wsi} \in \mathbb{R}^{C\times K}$ and $\mathbf{S_\text{tissue}^\text{patch} }\in \mathbb{R}^{N\times K}$, we then determine the contribution of each instance (tissue patch) within the whole slide image $X$ in classifying it into the respective whole slide classes. This allows us to aggregate class-specific visual features at the whole slide level
$\mathbf{F}_\text{wsi} \in \mathbb{R}^{d_{v}\times C}$ with image embedding dimension $d_{v}$ as follows:
\begin{equation}\label{eq:f_wsi}
    \mathbf{F}_\text{wsi} = \texttt{norm} \big(\mathbf{F} \times \mathbf{S}\big)
\end{equation}
where $\mathbf{F} = [\phi_\text{vision}(x_1), \hspace{0.5em} \dots \hspace{0.5em}, \phi_\text{vision}(x_N)]$, representing the array of the visual embedding of $N$ patches of a WSI extracted by $\phi_\text{vision}$. $\mathbf{S} = \mathbf{S_\text{tissue}^\text{patch}}\times {\mathbf{S}_\text{tissue} ^ \text{wsi}}'$ denotes the patch to slide correlation matrix, and \texttt{norm} normalizes the features to ensure they have a uniform magnitude.

\begin{figure}
    \centering
    \includegraphics[width=0.5\textwidth, height=7.3cm]{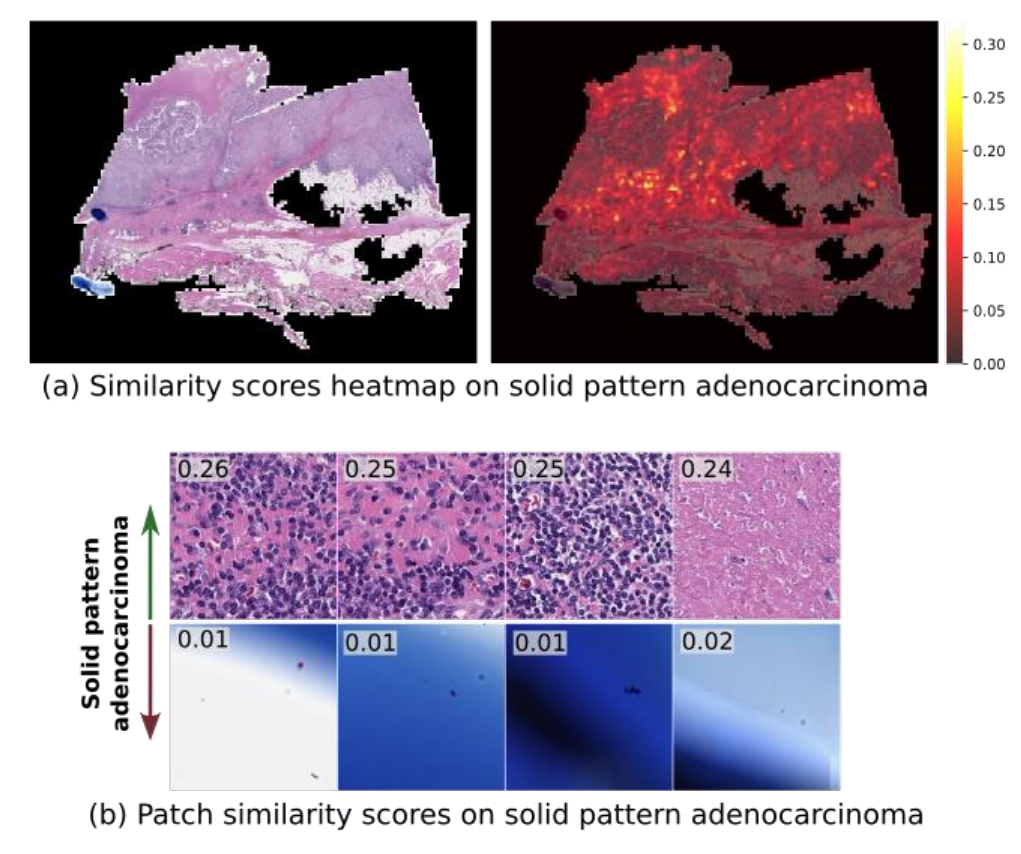}
    \caption{\textbf{SLIP similarity scores on Lung WSIs.} (a) Heatmaps for solid pattern adenocarcinoma. (b) Patches with highest (top) and lowest (bottom) similarity scores.} 
    \label{fig:attention}
\end{figure}
\noindent
{\bf Prompt-based fine-tuning.}
Prompt learning is highly effective for adapting pre-trained VLMs to new tasks with limited data, essential for large histology datasets where labeling is challenging. This approach enables few-shot learning, letting models work with minimal labeled examples, which is manageable for common cancer types. With fewer parameters to tune, prompt learning reduces computational costs compared to full fine-tuning, making it efficient for resource-limited settings. Inspired by \cite{zhou2022learning}, we append learnable prompts to textual slide class names, optimizing them during fine-tuning on a small set of labeled WSIs to better align with histopathological analysis. Leveraging the class-wise visual features $F_\text{wsi}$  from each WSI, we employ a supervised InfoNCE loss \cite{khosla2020supervised} to optimize the prompts as detailed in the following equation:
\begin{equation}
    \mathcal{L}_\text{InfoNCE}(X, c) = - \log \frac{\exp(F_\text{wsi}[:, c] \cdot \phi_\text{text}(y_\text{wsi}^c) / \tau)}{\sum_{i,j} \exp(F_\text{wsi}[:, i] \cdot \phi_\text{text}(y_\text{wsi}^j) / \tau)}
\end{equation}
where $c$ represents the index of ground truth whole slide label. The supervised InfoNCE loss is chosen to maximize mutual information between image and text embeddings, helping the model associate specific WSI visual features with correct textual annotations.

\noindent
\textbf{Model evaluation.} To evaluate the model’s performance, we first determine $F_\text{wsi}$ as the class-wise average visual embedding for the specified WSI image, as previously outlined in Equation \ref{eq:f_wsi}. The classification of a given WSI is then derived by identifying the class with the maximum alignment between its visual features and the corresponding textual prompt, following the criterion:
\begin{equation}
    c =  \argmax_j \hspace{0.5em} F_\text{wsi}[:, j] \cdot \phi_\text{text}(y_\text{wsi}^j)
\end{equation}

\section{Experiments}
\label{sec:experiments}
We evaluated our method on two H\&E-stained WSI datasets, chosen for fair comparisons (e.g., \cite{zhang2023text}).

\noindent
{\bf DHMC.} The Dartmouth lung cancer histology dataset \cite{wang2021transpath}\footnote{https://bmirds.github.io/LungCancer/} includes 143 FFPE WSIs annotated into Lepidic, Acinar, Papillary, Micropapillary, and Solid patterns. We focused on Lepidic, Acinar, and Solid subtypes, with a test split of [8, 26, 24] WSIs per class, generating 300×300 patches at 20× magnification and splitting WSIs 50/50 for training and validation.

\noindent
{\bf PatchGastric.} The stomach
tumor pathological image dataset \cite{tsuneki2022inference}\footnote{https://zenodo.org/records/6550925}, includes 262,777 patches (300×300) from 991 WSIs, divided into three gastric adenocarcinoma subtypes: Well-differentiated, Moderately differentiated, and Poorly differentiated, with a test split of [226, 212, 116] WSIs per class. Following \cite{zhang2023text}, WSIs were split 20/80 for training and validation.

\noindent
{\bf Few-shot setup.} 
We adapted pre-trained models by fine-tuning on target datasets, using 1-16 WSIs from PatchGastric and 1-8 from DHMC, selecting WSIs with the highest patch count per class. Patient-wise accuracy was averaged across classes, with zero-shot classification providing a baseline.

\noindent
\textbf{Implementation details.} Our approach builds on the ViT-B/16 \cite{radford2021learning} vision encoder from CLIP, extending to BiomedCLIP \cite{zhang2023large}, utilizing PubMedBERT for domain-specific text. We used 224×224 images, 16×16 patches, and an embedding dimension of $d_{v}=512$. All experiments ran on a single NVIDIA GeForce RTX 3080 GPU with  \(\tau\)= 0.01, learning rate 2e-4, SGD optimizer, and batch size of one.

\noindent
\textbf{Comparison against SOTA methods.} 
We benchmark our method against state-of-the-art (SOTA) MIL and VLM approaches, using CLIP and BiomedCLIP ViT-B/16 vision encoders for fair comparison. For MIL-based methods, we include CLAM \cite{lu2021data}, TransMIL \cite{shao2021transmil}, and PANTHER \cite{song2024morphological}, along with linear probing on a frozen encoder. For VLM approaches, we evaluate CITE \cite{zhang2023text}, CoOp \cite{zhou2022learning}, and TOP \cite{qu2024rise}, reproducing results with official code and settings.

On the PatchGastric classification task (\Cref{tab:merged_gastric_results}), SLIP outperformed all baselines, achieving 70.0\% accuracy in the "all shots" scenario with CLIP, bridging domain gaps. While CITE excelled in 1-shot and 2-shot, SLIP performed consistently across all shots, with a +14.1\% average gain over CoOp. Using BiomedCLIP, SLIP achieved high accuracy in 8-shot and "all shots," proving effective with complex histopathology images. Figure \ref{fig:dhmc} highlights SLIP's superior performance in limited-shot lung adenocarcinoma classification, with dual-similarity pooling and slide-level prompts enabling finer WSI analysis than conventional MIL methods.

\noindent
\textbf{Ablation on instance pooling.}
We perform an ablation study on instance pooling strategies, comparing our proposed pooling (SLIP-our pooling) with zero-shot (SLIP-zero), top-k pooling (SLIP-topk), and average pooling (SLIP-avg), as shown in~\Cref{tab:merged_gastric_results}. SLIP-topk, which selects top instances by relevance, achieves 66.7\% (CLIP, 16 shots) and 70.9\% (BiomedCLIP, all shots) but is generally outperformed by our method, which optimally aggregates features by focusing on slide-representative patches. Unlike average and top-k pooling, our approach selectively emphasizes patch-level features that best represent slide-level labels, significantly boosting performance in prompt-based few-shot adaptation.

\noindent
\textbf{Similarity scores visualization.}
Figure \ref{fig:attention} shows heatmaps and top patch similarity scores from our model on DHMC lung adenocarcinoma slides, focusing on solid patterns. Top panels (a) display attention heatmaps, with warmer colors highlighting dense tumor regions. Bottom panels (b) present patches with the highest and lowest similarity scores, capturing cohesive tumor areas validated by an expert pathologist. These visualizations demonstrate the model’s focus on key regions, enhancing subtype classification accuracy.

\noindent
\textbf{Ablation on tissue type prompts.} We conducted an ablation study on ChatGPT-generated tissue types on PatchGastric classification. Using 18 types improved 4-shot performance (62.2\%), while 10 types performed better in 16-shot settings (67.9\%), highlighting a trade-off between context richness and relevance in WSI classification.

\section{Conclusion}
\label{sec:conclusion}
Our paper presents a novel few-shot MIL approach for histopathology, leveraging VLMs to improve WSI classification accuracy. Through prompt learning and a dual-similarity pooling framework, our model aligns visual features with tissue types and class descriptors, surpassing traditional MIL and VLM methods and enhancing interpretability and generalizability. Future work will target computational efficiency and address potential biases in few-shot learning.

\section{Compliance with Ethical Standards}
This research study used human subject data from publicly available datasets \cite{wang2021transpath,tsuneki2022inference}. Ethical approval was not required, as confirmed by the license attached and its publication description. The dataset was de-identified by the original providers to ensure privacy and confidentiality.

\bibliographystyle{IEEEbib}
\bibliography{egbib}

\end{document}